\documentclass[final]{cvpr}

\usepackage{amsmath,amsfonts,bm}

\def\eqref#1{equation~\ref{#1}}

\def\1{\bm{1}}

\DeclareMathAlphabet{\mathsfit}{\encodingdefault}{\sfdefault}{m}{sl}
\SetMathAlphabet{\mathsfit}{bold}{\encodingdefault}{\sfdefault}{bx}{n}

\usepackage{times}
\usepackage{epsfig}
\usepackage{graphicx}
\usepackage{url}
\usepackage{microtype}
\usepackage{graphicx}
\usepackage{subfigure}
\usepackage{booktabs} 
\usepackage{xcolor}
\usepackage{amsmath, amssymb, amsthm}
\usepackage{multirow}
\usepackage{diagbox}

\usepackage{booktabs}
\usepackage{float}
\usepackage{hhline}
\usepackage{tabularx}
\newcolumntype{Y}{>{\centering\arraybackslash}X}
\newcolumntype{s}{>{\hsize=.8\hsize}Y}
\newcolumntype{t}{>{\hsize=.6\hsize}Y}

\newcolumntype{?}{!{\vrule width 1pt}}
\usepackage{multirow}
\usepackage{makecell}
\usepackage{stfloats}
\usepackage{wrapfig}
\usepackage{dsfont}
\usepackage{nopageno}

\usepackage[pagebackref=true,breaklinks=true,colorlinks,bookmarks=false]{hyperref}

\newcommand{\minisection}[1]{\vspace{0mm}\noindent{\textbf{#1}.}}

\begin{document}

\title{Permuted AdaIN: Reducing the Bias Towards \\ Global Statistics in Image Classification}

\author{Oren Nuriel\quad Sagie Benaim \quad Lior Wolf \\
Tel Aviv University}

\maketitle

\begin{abstract}

Recent work has shown that convolutional neural network classifiers overly rely on texture at the expense of shape cues. We make a similar but different distinction between shape and local image cues, on the one hand, and global image statistics, on the other. Our method, called Permuted Adaptive Instance Normalization (pAdaIN), reduces the representation of global statistics in the hidden layers of image classifiers. pAdaIN samples a random permutation $\pi$ that rearranges the samples in a given batch. Adaptive Instance Normalization (AdaIN) is then applied between the activations of each (non-permuted) sample $i$ and the corresponding activations of the sample $\pi(i)$, thus swapping statistics between the samples of the batch. Since the global image statistics are distorted, this swapping procedure causes the network to rely on cues, such as shape or texture. By choosing the random permutation with probability $p$ and the identity permutation otherwise, one can control the effect's strength. 

With the correct choice of $p$, fixed apriori for all experiments and selected without considering test data, our method consistently outperforms baselines in multiple settings. In image classification, our method improves on both CIFAR100 and ImageNet using multiple architectures. In the setting of robustness, our method improves on both ImageNet-C and  Cifar-100-C for multiple architectures. In the setting of domain adaptation and domain generalization, our method achieves state of the art results on the transfer learning task from GTAV to Cityscapes and on the PACS benchmark.

\end{abstract}

\section{Introduction}
One of the early successes of computer vision was a face recognition system by Sakai et al.~\cite{Sakai-1971-15077} that employed a simple neural network classifier. As it turns out, the network was relying on global image statistics, namely the average brightness, to perform recognition.

In this work, we demonstrate that removing the reliance on global image statistics improves classification results in modern networks. To mitigate the effect of the global statistics,  a deliberate mismatch between the activations of a layer and its accumulated statistics is created.  By normalizing with unmatched statistics, the distribution of activation values becomes unreliable as a source for label information. While changing the global statistics of an image to another was explored in the context of generation \cite{gatys2015texture, huang2017arbitrary}, we show that it is also useful in a variety of discriminative settings.

Our work is similar in spirit but different in conclusion from recent work \cite{carlucci2019domain, geirhos2018imagenet, hermann2019exploring, shi2020informative, wang2019learning, zhang2019interpreting} that has identified a bias toward texture at the expense of shape.
Such recent methods can often improve the performance of the image classifier on the test set and have been shown to dramatically increase the accuracy of the classifier on shifted image domains, in which image transformations change the image statistics but leave most of the shape unchanged.

In our work, we also show classification and domain generalization improvements. 
However, we demonstrate that the increase in classification performance occurs simultaneously for both category-based image recognition and texture recognition. This suggests that while the texture is often defined as local image statistics, becoming invariant to global image statistics improves both shape and texture recognition.

We demonstrate the effectiveness of our method in a number of settings. First, we demonstrate how classification performance improves when adding our permutation-based regularization. Our method improves accuracy on both CIFAR100 and ImageNet on multiple architectures trained in a vanilla fashion.
Second, we train a linear classifier on top of a pre-trained image-classification network's representation layer and show that the accuracy of texture classification peaks exactly when the image classification results are maximized. We show that when this happens, the network's representation of shape does not deteriorate. 
Next, we demonstrate that our method can reduce the adverse effect of domain shift, by testing it in the setting of domain adaptation and more broadly in domain generalization. Our method achieves state of the art results on the domain adaptation from GTA5 to Cityscapes semantic segmentation and on the PACS dataset. 
Lastly, we show that our method allows for a greater robustness when handling corrupted images, where our method is superior to all baseline methods.
In the setting of robustness, our method improves on both ImageNet-C and Cifar-100-C for multiple architectures.

\section{Related Work}

\minisection{Bias towards texture} A large body of work has shown that, unlike humans, networks tend to be biased towards textures in making decisions. Gatys et al.,~\cite{gatys2015texture} have shown that training a linear classifier on top of a VGG19 texture representation achieves similar performance to training VGG19 directly on this task.  Geirhos et al.,~\cite{geirhos2018imagenet} observed this phenomenon in the context of pretrained ImageNet CNNs. They presented the  `Stylized-ImageNet’ dataset, which is a version of ImageNet where the image style is altered, and show that training with this dataset forces the network to learn a shape-based representation. Hermann and Kornblith~\cite{hermann2019exploring} explored the role of different factors, such as the training objective or architecture, on reducing texture bias. 
 
However, this resulted in a degradation in the network's performance. Unlike these methods, from the technical perspective, our method does not rely on the additional supervision in the form of extended or modified datasets and instead directly modifies the architecture of the network. Our interpretation of the results is also different. We show that      manipulating the global statistics, which are directly linked to style, does not hurt texture recognition.

Several contributions attempt to alleviate texture bias, by proposing an architectural change or a new training objective. Shi et al.,~\cite{shi2020informative} develop a Dropout-like algorithm. 
Wang et al.,~\cite{wang2019learning} penalize shallow layers for having predictive power. 
Zhang and Zhu~\cite{zhang2019interpreting} show that adversarial training reduces texture bias. Carlucci et al.,~\cite{carlucci2019domain} propose to reduce texture bias, by training the network to solve jigsaw puzzles. Unlike these methods, our method makes use of a novel normalization layer, which, as shown in Sec.~\ref{sec:motivation},  directly affects the dependence on global image statistics.

\smallskip
\minisection{Normalization and style transfer}
Batch Norm~\cite{ioffe2015batch} has become a standard mechanism for effectively training deep neural networks by normalizing activations by the statistics of the minibatch. To reduce minibatch dependencies, several alternatives were proposed, including Layer Normalization~\cite{ba2016layer}, Instance Normalization~\cite{ulyanov2016instance}, and Group Normalization~\cite{wu2018group}. Our work utilizes the ability to swap the style statistics of images as part of a novel normalization layer. Unlike our normalization layer, its role is not to support efficient training, but to direct the network toward the desired emphasis on shape and fine details.

Instance norm by Ulyanov  et al.,~\cite{ulyanov2016instance}, can be seen as a form of style normalization by normalizing feature statistics. Building on this view, Huang  and  Belongie~\cite{huang2017arbitrary} proposed Adaptive Instance Normalization (AdaIN) as a form of style transfer, by first normalizing the target image style statistics and then rescaling by source image style statistics.
These style manipulations through normalization layer methods are mostly applied in the generative setting, where they can be used for texture synthesis and style transfer, while our method focuses on image recognition.

Our method builds upon AdaIN to swap the style statistics of different element activation. As far as we are aware, while other methods use style transfer to construct an improved dataset~\cite{hermann2019exploring, geirhos2018imagenet} to alleviate the reliance on global image statistics, our method is the first to do so within a network.

\section{Method}

We are interested in the distinction between global image statistics on the one hand, and other global cues such as shape as well as local cues, on the other. 
Global statistics are statistics (such as mean and standard deviation) measured from all the pixels of the image. These include, for example, brightness, contrast, lightning and global color changes. Changing the style of the image typically changes such statistics. Global cues refer to any cues present in the entire image or large patches of it, but not in small patches. These include global statistics but also shape information (such as an edge map of a cat). Local cues refer to any information present in small patches in the image, which may include both texture and shape information within those patches.

\subsection{Motivation}
\label{sec:motivation}

To motivate our method, we conduct a simple experiment visualizing the effect of swapping statistics of intermediate layer representations of an autoencoder. The autoencoder was trained to minimize the reconstruction error on the Stanford Car Dataset~\cite{KrauseStarkDengFei-Fei_3DRR2013}. We consider two image inputs, $a$ and $b$ and inspect the effect on the reconstruction of $a$, when swapping their statistics at different layers of the pretrained encoder. The decoder is left unchanged. As a baseline method, we also observe the effect of transferring the style of image $b$ to $a$ using the method of Gatys et al.,~\cite{gatys2015texture}.

\begin{figure*}
\centering
\begin{tabular}{ccccccccc}
\includegraphics[width=0.087\linewidth]{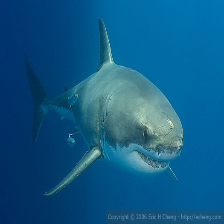} &
\includegraphics[width=0.087\linewidth]{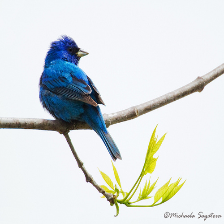} &
\includegraphics[width=0.087\linewidth]{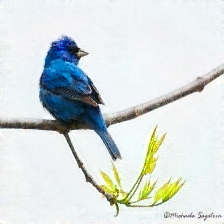} &
\includegraphics[width=0.087\linewidth]{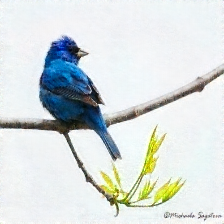} &
\includegraphics[width=0.087\linewidth]{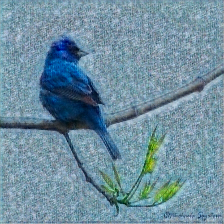} &
\includegraphics[width=0.087\linewidth]{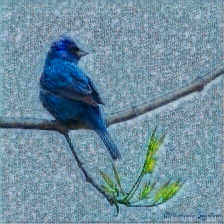} &
\includegraphics[width=0.087\linewidth]{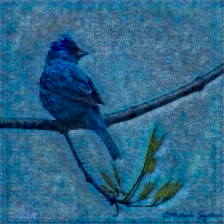} &
\includegraphics[width=0.087\linewidth]{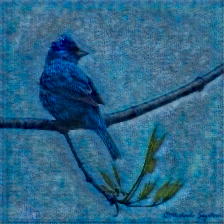}  &
\includegraphics[width=0.087\linewidth]{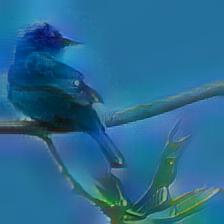} \\
\includegraphics[width=0.087\linewidth]{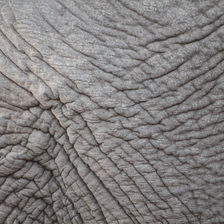} &
\includegraphics[width=0.087\linewidth]{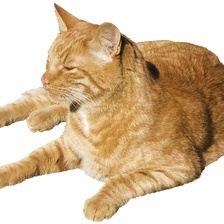} &
\includegraphics[width=0.087\linewidth]{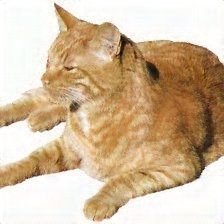} &
\includegraphics[width=0.087\linewidth]{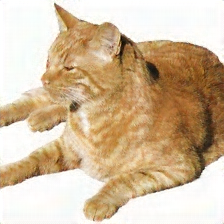} &
\includegraphics[width=0.087\linewidth]{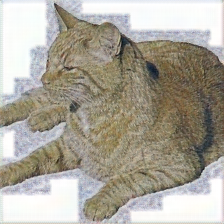} &
\includegraphics[width=0.087\linewidth]{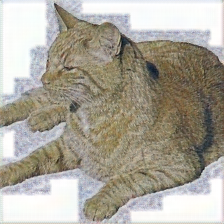} &
\includegraphics[width=0.087\linewidth]{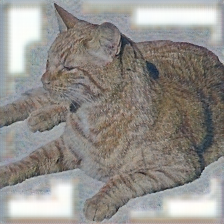} &
\includegraphics[width=0.087\linewidth]{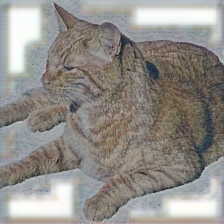} &
\includegraphics[width=0.087\linewidth]{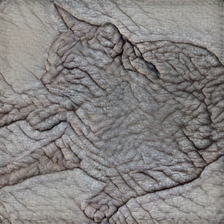} \\
Image b& Image a & Recon a & L. 0 & L. 0-1 & L. 0-2 & L. 0-3 & L. 0-4 & \cite{gatys2015texture}\\
\end{tabular}
\vspace{0.1cm}
\caption{Applying pAdaIN at inference on different layers of an encoder trained as part of an auto-encoder. The input and reconstructed images are shown on the left. The reconstructed results when applying pAdaIN on different layers of the encoder are shown subsequently. The last image on the right is the result of applying style transfer using the method of Gatys et al.,~\cite{gatys2015texture}. L.=layer; Recon=Reconstruction.} 
\label{fig:motivation2}
\vspace{-0.4cm}
\end{figure*}

As can be seen in Fig.~\ref{fig:motivation2}, when swapping the image statistics used by the AdaIN module, the reconstructed image has similar global statistics, such as color and overall image appearance of $b$ but the finer details of image $a$ are preserved. Applying this swapping on more layers, results in a larger transfer of the global statistics of $b$. 
In contrast, when using style transfer, the fine details of $a$ are borrowed from image $b$ and are no longer preserved. For example in Fig.~\ref{fig:motivation2}, when applying style transfer, the bird on a tree was given the fine details of the shark under water, and similarly a cat was given the texture of the elephant skin. Such details were not transformed by swapping the normalization parameters. We argue that preserving the fine details, while transferring the global ones, results in an augmented sample that can be utilized to improve classification accuracy and make the trained network more robust to imaging conditions and better suited for generalization to new visual domains.

\subsection{Adaptive Instance Normalization}

We begin by defining Instance Normalization (IN), as formulated in \cite{ulyanov2016instance} and \cite{huang2017arbitrary}.
For a given convolutional neural network, let the output activations of a given convolutional layer be $x \in \mathbb{R}^{N \times C \times H \times W}$, where $N$ is the batch size, $C$ the number of channels, $H$ the height of the layer, and $W$ its width. Instance Norm is then defined as:
\begin{equation}
\textrm{IN}(x)= \gamma\left(\frac{x-\mu(x)}{\sigma(x)}\right)+\beta\,,
\end{equation}
where $\mu(x)$ and $\sigma(x)$, both in $\mathbb{R}^{N\times C}$, are the mean and standard deviation, computed along the spatial dimensions ($H\times W$) for each channel ($c$) and sample in the batch ($n$):
\begin{equation}
\mu_{nc}(x) = \frac{1}{HW}\sum_{h=1}^{H}\sum_{w=1}^{W}x_{nchw}
\end{equation}
\begin{equation}
\sigma_{nc}(x) = \sqrt{\frac{1}{HW}\sum_{h=1}^{H}\sum_{w=1}^{W}(x_{nchw} - \mu_{nc}(x))^{2} + \epsilon}\,.
\end{equation}
$\gamma$ and $\beta$, both in $\mathbb{R}^{N\times C}$, are the re-scaling affine  parameters learned independently of $x$.

The above operation is applied in the same manner both at train and test time. As detailed in \cite{huang2017arbitrary}, IN can be viewed as normalizing the style statistics of each input in the batch.
Adaptive Instance Normalization (AdaIN) builds upon this view, by first normalizing the style statistics of an input $a$, thus extracting its content, and then scaling the normalized output by the statistics of a target style input $b$. This allows the transfer of style from $b$ to $a$. Specifically, let $a, b \in \mathbb{R}^{C \times H \times W}$, then AdaIN is defined as:

\begin{equation}
\label{eq:adain}
\textrm{AdaIN}(a, b)= \sigma(b)\left(\frac{a-\mu(a)}{\sigma(a)}\right)+\mu(b)
\end{equation}
where $\mu(a)$ and $\sigma(a)$ (resp. $\mu(b)$ and $\sigma(b)$) are the mean and standard deviation of $a$ (resp. $b$) over its spatial dimension,  computed for each channel.

\subsection{Permuted AdaIN}

Given an input activations map $x \in \mathbb{R}^{N \times C \times H \times W}$, let $\pi(x) = [x_{\pi(1)}, x_{\pi(2)}, \dots, x_{\pi(N)}]\in \mathbb{R}^{N\times C\times H \times W}$ be the result of applying a permutation $\pi$ to the elements of a given mini-batch $x = x_1, \dots, x_N$ along the minibatch axis.

The result of applying pAdaIN on a single sample $x_i$ in the context of its batch and for a given permutation $\pi$ is:
\begin{equation}
\textrm{p-IN}^\pi(x_i) = \textrm{AdaIN}(x_i, x_{\pi(i)})\,.
\end{equation}

$\textrm{pAdaIN}$ is then defined for the entire tensor $x$:
      \begin{equation*}
    \textrm{pAdaIN}(x) = \\
    \begin{cases}
      x, & \text{probability }\ $p$ \\
      (\textrm{p-IN}^\pi(x_1), .., \textrm{p-IN}^\pi(x_N)) & \text{otherwise}
    \end{cases}
  \end{equation*}
where $\pi$ is a uniformly chosen permutation, and $p$ is a hyperparameter fixed ahead of training.
pAdaIN is only applied during training time and not at test time. We apply pAdaIN to the output activations of all convolutional layers and in particular, before applying batch normalization.

Backpropagation is applied through $x$ but not through $\pi(x)$. Specifically setting $a=x_i$ and $b=x_{\pi(i)}$ in Eq.~\ref{eq:adain}, we regard $\mu(x_{\pi(i)})$ and $\sigma(x_{\pi(i)})$ as constant and do not backpropagate through them. Performing a different update, such as one on $\mu(x_{\pi(i)})$ and $\sigma(x_{\pi(i)})$ leads to sub-optimal results, as shown in Sec.~\ref{sec:ablation}. Mixing batch information in the forward pass during training is used as a regularization to the model, and is shown to improve generalization (see Sec.~\ref{sec:generalization}). Backpropagating gradients on both $x$ and $\pi(x)$ causes the loss on a sample $x_i$ of the batch to affect the gradients of another sample $x_{\pi(i)}$ in the batch,  which is undesired.

\noindent\textbf{Effect of Batch Norm.}\quad Batch Norm ($BN$) normalizes channel-wise statistics yet does not undo the effect of our method. To see this, we first define the $BN$ operation: 
\begin{equation}
\mu_{c}(x) = \frac{1}{N}\sum_{n=1}^{N}\mu_{nc}(x)
\end{equation}
\begin{equation}
\sigma_{c}(x) = \sqrt{\frac{1}{NHW}\sum_{n=1}^{N}\sum_{h=1}^{H}\sum_{w=1}^{W}(x_{nchw} - \mu_{c}(x))^{2} + \epsilon}
\end{equation}
\begin{equation}
BN(x) = \gamma\left(\frac{x-\mu_c(x)}{\sigma_c(x)}\right)+\beta\,.
\end{equation}
For some parameters $\gamma$ and $\beta$. After applying $BN$, we have:
\begin{equation}
    \mu_{nc}(BN(x))=\frac{\gamma}{\sigma_c(x)} \cdot (\mu_{nc}(x) - \mu_c(x)) + \beta \,,
\end{equation}
\begin{equation}
    \sigma_{nc}(BN(x))=\frac{\gamma \cdot \sigma_{nc}(x)}{\sigma_c(x)}
\end{equation}
With pAdaIN, when statistics are swapped (i.e not identity):
\begin{equation}
    \label{eq:bn_mu}
    \mu_{nc}(BN(\textrm{pAdaIN}(x)))=\frac{\gamma}{\sigma_c} \cdot (\mu_{\pi(n)c}(x) - \mu_c(x)) + \beta\,,
\end{equation}
\begin{equation}
    \label{eq:bn_sigma}
    \sigma_{nc}(BN(\textrm{pAdaIN}(x)))=\frac{\gamma \cdot \sigma_{\pi(n)c}(x)}{\sigma_c(x)}
\end{equation}
Eq.~\ref{eq:bn_mu} and Eq.~\ref{eq:bn_sigma} follow from pAdaIN shifting channel-wise statistics. 
The statistics of channel $c$, for sample $n$, after applying pAdaIN, are the same as that of sample $\pi(n)$, beforehand.
Hence $BN$ does not undo the swapping of statistics, but rather scales them by batch-wise statistics.

\section{Experiments}

Our experiments explore classification accuracy of both objects and texture, robustness to image corruption, and generalization to new domains. Unless otherwise mentioned, pAdaIN is applied with a fixed choice of $p=0.01$.

\subsection{Image Classification}
\label{sec:classification}

We  evaluate pAdaIN in the context of image classification on both CIFAR100 and ImageNet. To evaluate pAdaIN, for every architecture, we add a pAdaIN layer before every use of batch normalization and after using a convolutional layer.

For CIFAR100, we consider the architectures of VGG19\cite{simonyan2014very}, InceptionV4\cite{szegedy2017inception}, PyramidNet \cite{han2017deep}, ResNet18 and ResNet50 \cite{he2016deep}.
During training, we apply a padding of $4$, a random crop and a random rotation of up to $15\%$, resulting in images of size $32 \times 32$. The networks are trained on a batch size of $128$ with an SGD  with a momentum of $0.9$ and a weight decay of $5e^{-4}$. 
We use $200$ epochs, start training with a learning rate of $0.1$ and divide the learning rate by $5$ at epochs $60$, $120$ and $160$.
For ImageNet, we consider the architectures of ReseNet50, ResNet101 and ResNet152 \cite{he2016deep}.
We train for $300$ epochs, and use standard augmentations of resizing to $256 \times 256$ and applying a random crop of $224 \times 224$ and then applying a random horizontal flip. The learning rate is initiated to $0.1$ for ResNet50, ResNet101, and ResNet152, after which it is reduced by a factor of $10$ every $75$ epochs. SGD with momentum is used as the optimizer. The batch size, weight decay and momentum were set to $256$, $1e^{-4}$ and $0.9$ respectively.

For all experiments, a default value of $p=0.01$ is used. In Tab.~\ref{tab:cifar_classification} and Tab.~\ref{tab:classification_imagenet} we compare, for the different architectures, the result of training the network with pAdaIN and without pAdaIN (Baseline). Other than the use of pAdaIN, which does not add any learnable parameters to the network, the same architecture and training procedure is used.  As can be seen, our method outperforms the baseline on the above datasets. The improvement is consistent across networks with a vastly different number of parameters, such as ResNet18 and ResNet50 for CIFAR100 and ResNet50, ResNet101, and ResNet152 for ImageNet. The improvement is also consistent across different model types, such as VGG, Inception, PyramidNet, and ResNet for CIFAR100. 

\begin{figure}[t]
 \centering
	\includegraphics[width=0.85\columnwidth]{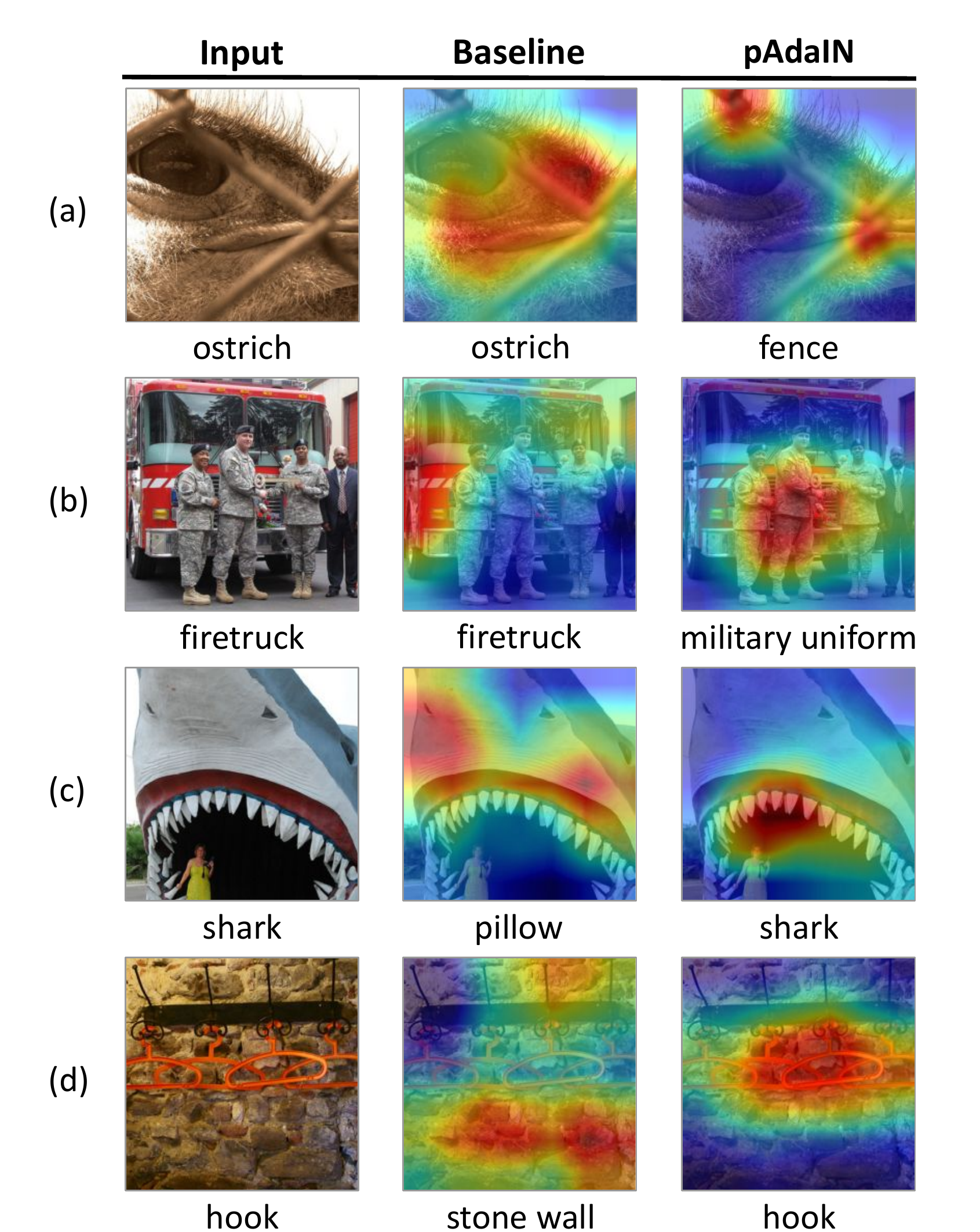}
	\caption{GradCam~\cite{selvaraju2017grad} visualizations and predictions for ImageNet trained ResNet50 models, with and without pAdaIN. Ground truth labels and model predictions appear on the bottom of each image respectively.}
	\vspace{-0.5cm}
	\label{fig:gradcam}
\end{figure}

We consider the effect of changing $p$ on the overall accuracy. This is done for the ResNet18 and ResNet50 models trained on CIFAR100 and ImageNet, respectively. As can be seen in Fig.~\ref{fig:texture_accuracy}(d), increasing the value of $p$ up to $0.01$ results in improved accuracy, after-which accuracy drops.

Lastly, to qualitatively analyze our method, we considered the two ResNet50 models trained on ImageNet, either with or without pAdaIN.
Fig.~\ref{fig:gradcam} depicts four examples along with the GradCAM~\cite{selvaraju2017grad} visualization for the predicted class
pAdaIN concentrates on the foreground and so, for (a), predicts a chainlink fence. The vanilla (baseline) model predicts an ostrich, as it relies more on global statistics.
While both answers could be correct, the GT (ground truth) annotation is that of an ostrich and so this is regarded as an error of pAdaIN. 
In contrast, (c) depicts an image of shark on land. Our model relies less on global context (such as the shark being at sea) and so predicts a shark (which corresponds to the GT annotation). The vanilla model predicts a pillow. Similarly, for (b) (resp. (d)) image, pAdaIN focuses on persons with military uniform (resp. hook), and the vanilla model on the firetruck (resp. stone wall) in the background.

\begin{table}
\centering
\begin{tabular}{l|c|c}
\toprule
Architecture   
 & Baseline & pAdaIN \\ 
\midrule
VGG19 & 72.30 & \textbf{72.90} \\

ResNet18 & 76.13 & \textbf{77.82}\\

ResNet50 & 78.22 & \textbf{79.03} \\

InceptionV4 & 78.00 & \textbf{79.50}\\

PyramidNet & 83.49 & \textbf{84.17} \\
\bottomrule
\end{tabular}
\smallskip
\caption{Top-1 accuracy for CIFAR100 on different architectures. }
\label{tab:cifar_classification}
\vspace{-0.2cm}
\end{table}

\begin{table}
\centering
\begin{tabular}{l|cc|cc}
\toprule
\multirow{2}{*}{Architecture} & \multicolumn{2}{c|}{Top-1} & \multicolumn{2}{c}{Top-5}\\
  & Baseline & pAdaIN & Baseline & pAdaIN \\
\midrule
ResNet50 & 77.1 & \textbf{77.7} &  93.63 & \textbf{93.93} \\
ResNet101 & 78.13 & \textbf{78.8} & 93.71 &  \textbf{94.35}  \\
ResNet152 & 78.31 & \textbf{79.13} & 94.06 & \textbf{94.64}   \\
\bottomrule
\end{tabular}
\smallskip
\caption{Top-1/Top-5 ImageNet accuracy on different architectures.  }
\label{tab:classification_imagenet}
\vspace{-0.4cm}
\end{table}

\subsection{Texture and Shape Representation}

\begin{figure*}
\centering
\begin{tabular}{cccc}
\includegraphics[width=0.22\linewidth]{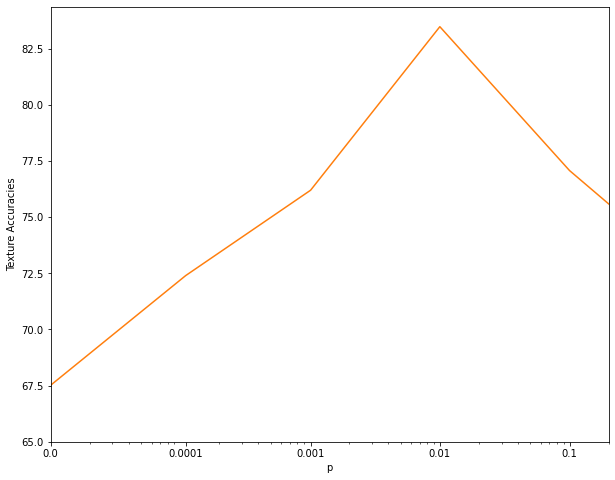} &
\includegraphics[width=0.22\linewidth]{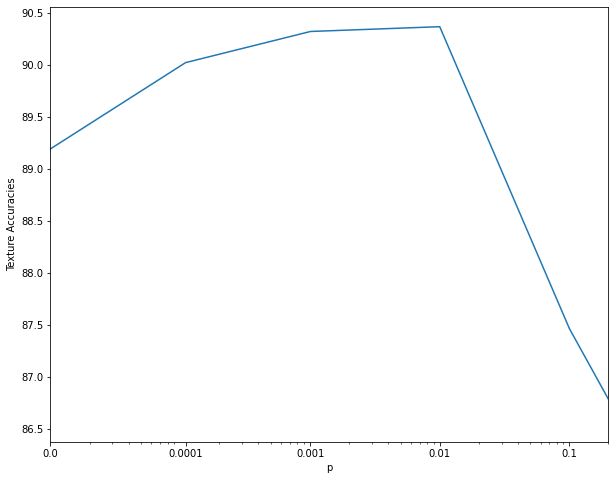} &
\includegraphics[width=0.22\linewidth]{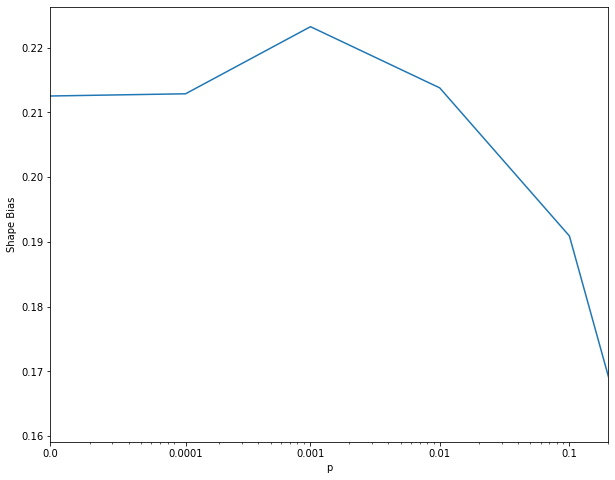} &
\includegraphics[width=0.22\linewidth]{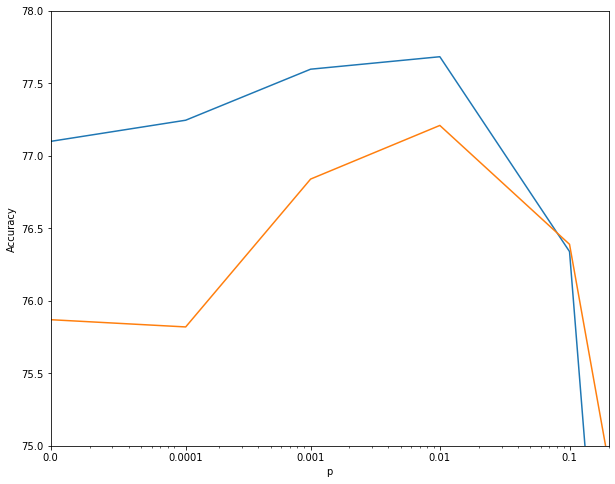} \\
(a) & (b) & (c) & (d) \\
\end{tabular}
\caption{(a) Texture accuracy for different values of $p$ for a ResNet18 model trained on CIFAR100. (b) As in (a), but for ResNet50 model trained on ImageNet. (c) Shape bias for a ResNet50 model trained on ImageNet, for a range of values of $p$. (d) Accuracy for models trained with pAdaIN for various values of $p$. In blue, a ResNet50 model trained on ImageNet and in orange, a ResNet18 model trained CIFAR100. For $p$ values above $0.1$ (not shown), accuracy drops significantly below $75\%$ for both ImageNet and CIFAR100.}
\label{fig:texture_accuracy}
\vspace{-0.3cm}
\end{figure*}

To evaluate the resulting feature representation for texture recognition, the texture surface dataset~\cite{huang2020compact} was employed. It consists of $64$ classes, with a total of $8674$ images.  
Since some textures can be similar to others we sample $10$\% for training and use the rest for test. 
 
Our training procedure consists of freezing the backbone of the trained model, and training a linear classifier on top of the last representation layer {of both ResNet models} (before the label logits) to correctly classify the texture class. A high accuracy indicates that the model captures texture more strongly in its representation layer.

For ImageNet and CIFAR100, we consider the texture accuracy when training with pAdaIN for various values of $p$. As can be seen in Fig.~\ref{fig:texture_accuracy} (a,b), a value of $p=0.01$ results in the best performing model.
As pAdaIN is applied at each layer with probability $p$ independently, setting $p$ too high can result in an excessive change of statistics, thus resulting in degradation in accuracy.
We note that this coincides in the value of $p$ having the best overall accuracy for both ImageNet and CIFAR100, as shown in Fig.~\ref{fig:texture_accuracy}(d). 

Mostly, an increase (resp. decrease) in the value of $p$ results in an increase (resp. decrease) of both overall accuracy and texture accuracy. To evaluate this connection in the context of previous work that aimed at eliminating texture bias, we repeat the experiment with such a method. 

Specifically, we consider a model trained as described in Geirhos et al.,~\cite{geirhos2018imagenet} on a combination of ImageNet and a stylized ImageNet. We measure its texture accuracy, as above. The ImageNet classification accuracy of the employed Shape-ResNet increases from $76.13$ to $76.72$. Concurrently, the texture accuracy drops from $89.2\%$ to $88.7\%$.
This indicates that unlike our method which preserves local cues, Geirhos et al.,~\cite{geirhos2018imagenet} do not. Their increase in performance is due to the increased utilization of global cues at the expense of local ones. Our method improves the accuracy without reducing the recognizability of local textures.

Furthermore, we demonstrate that, for $p\leq0.01$, 
while the representation of global statistics, such as background color is distorted through the use of pAdaIN, the representation of shape is not. 
To show this, we consider the shape bias measure of Geirhos et al., \cite{geirhos2018imagenet} on the cue conflict dataset. 

This dataset was crafted to evaluate the shape bias of an ImageNet trained model and is composed of $1280$ images. Each image has two labels: a texture label and a shape label. The texture and shape labels are taken from $16$ different classes. Each image is the product of performing iterative style transfer \cite{gatys2016image} between an image from a texture dataset, containing the texture corresponding to one of the texture classes, and a natural colored image of an object with a white background from one of the shape classes. An example can be seen with the elephant skin (texture) and the cat (shape) adopted from \cite{geirhos2018imagenet} in Fig.~\ref{fig:motivation2}. A correct prediction is considered a prediction that matches one of the two classes that compose a test image, i.e., either the shape class or the texture class. Given an ImageNet trained model, the shape bias is computed as the proportion of correct shape predictions which the model makes out of all the correct predictions (either correct texture or shape). 

\begin{table*}
\resizebox{\textwidth}{!}{
\renewcommand{\arraystretch}{1.2}
    \begin{tabular}{l | c c c c c c c c c c c c c c c c c c c | c }

 \hline

 Method & Road & SW & Build & Wall & Fence & Pole & TL & TS & Veg. & Terrain & Sky & PR & Rider & Car & Truck & Bus & Train & Motor & Bike & mIoU\\
 \hline

 Source only  &57.9&17.4&71.5&19.3&18.3&25.39&32.5&16.8&82.3&28.2&78.0&55.3&31.3&71.6&19.1&26.8&9.2&26.3&13.7&37.0\\
Source only + pAdaIN &57.2&20.2&71.6&28.3&19.1&26.1&33.6&13.0&82.1&29.0&69.5&56.7&33.0&67.5&27.8&35.1&\textbf{17.6}&33.7&14.5&38.7\\
\cline{1-21}
AdaptSegNet \cite{Tsai_adaptseg_2018} & 86.5& 36.0& 79.9& 23.4& 23.3& 23.9& 35.2& 14.8& 83.4& 33.3& 75.6& 58.5& 27.6& 73.7& 32.5& 35.4& 3.9& 30.1& 28.1& 42.4\\
SIBAN \cite{Luo_2019_ICCV}&88.5 &35.4 &79.5& 26.3& 24.3& 28.5& 32.5& 18.3& 81.2& 40.0& 76.5& 58.1& 25.8& 82.6& 30.3& 34.4& 3.4& 21.6& 21.5& 42.6\\
CLAN \cite{Yawei2019Taking} & 87.0 &27.1& 79.6& 27.3& 23.3& 28.3& \textbf{35.5}& 24.2& 83.6& 27.4& 74.2& 58.6& 28.0& 76.2& 33.1& 36.7& 6.7& 31.9& 31.4& 43.2\\
AdaptPatch \cite{Tsai_DA4Seg_ICCV19} &92.3 &51.9& 82.1& 29.2& 25.1& 24.5& 33.8& \textbf{33.0}& 82.4& 32.8& 82.2& 58.6& 27.2& 84.3& 33.4& 46.3& 2.2& 29.5& 32.3& 46.5  \\
ADVENT \cite{vu2018advent}&89.4 &33.1 &81.0 &26.6 &26.8 &27.2 &33.5 &24.7 &83.9 &36.7 &78.8 &58.7 &30.5 &84.8 &38.5 &44.5 &1.7 &31.6& 32.4 &45.5\\
FADA \cite{Wang2020FADA}& 92.5&47.5&85.1&37.6&\textbf{32.8}&\textbf{33.4}&33.8&18.4&85.3&37.7&83.5&63.2&\textbf{39.7}&87.5&32.9&47.8&1.6&34.9&\textbf{39.5}&49.2\\
FADA \cite{Wang2020FADA} + pAdaIN & \textbf{93.3}&\textbf{55.7}&\textbf{85.6}&\textbf{38.3}&29.6&31.2&34.2&17.8&\textbf{86.2}&\textbf{41.0}&\textbf{88.8}&\textbf{65.1}&37.1&\textbf{87.6}&\textbf{45.9}&\textbf{55.1}&15.1&\textbf{39.4}&31.1&\textbf{51.5}\\
 \hline
\end{tabular}}
\smallskip
\caption{Experimental results for unsupervised domain adaptation and domain generalization (source only) on GTA5 $\to$ Cityscapes in the task of semantic segmentation, using DeepLabv2~\cite{chen2017deeplab} with ResNet101 backbone architecture.}
\label{tab:gta5_results}
\vspace{-0.3cm}
\end{table*}

As can be seen in Fig.~\ref{fig:texture_accuracy}(c), increasing the value of $p$ up to $0.01$ does not degrade the shape bias beyond that of a model with no pAdaIN ($p=0$), and, in fact, increases it slightly for $p=0.001$. At $p=0.01$, both the texture recognition ability (Fig.~\ref{fig:texture_accuracy}(b)) and the accuracy (Fig.~\ref{fig:texture_accuracy}(d)) are maximal. 
Evidently, while our model  strengthens local cues  as it improves classification, its affinity toward shape-based classification is not reduced. 
For $p>0.01$, we notice a decrease in shape bias and of accuracy. This indicates that using a value of $p$ which is too high may adversely affect the shape representation and subsequently the model's accuracy.

To further validate that pAdaIN does not support texture classification at the expense of shape, we evaluate our method on two shape oriented datasets. The first is ImageNet-Sketch~\cite{wang2019learning}, which consists of $50000$ sketch-like images, with $50$ images for each of the $1000$ ImageNet classes. The second is the Edges dataset~\cite{geirhos2018imagenet}, which consists of $160$ images of $16$ different objects with a white background, processed by the Canny edge extractor \cite{canny1986}. 

An ImageNet trained model with pAdaIN ($p=0.01$) (no finetuning), achieves $26.0\%$ and $26.9\%$ accuracy on ImageNet-sketch and the Edges dataset, respectively. In contrast, an ImageNet trained model without pAdaIN achieves a lower accuracy of $24.5\%$ and $24.4\%$, respectively.

\subsection{Domain Adaptation}
\label{sec:domain_adaptation}

Of particular interest is the ability of image classifiers to generalize in the settings where the test distribution is shifted compared to  the train distribution. In the setting of \textit{domain adaption}, one is given a labeled source data and an unlabeled target data and is asked to generalize well on both the source and target distributions~\cite{DBLP:conf/alt/Mansour09,Ganin:2016:DTN:2946645.2946704}.

We evaluate our method on the pixel-wise classification task of semantic segmentation,  in the setting of domain adaption. At train time, we are given access to training images from both the source and the target domain. However, the semantic segmentation labels are only available for the images from the source domain.  The objective is to maximize performance on the target domain.

We consider the state of the art method of Wang et al.,~\cite{Wang2020FADA} for which a three-step approach is undertaken, and apply pAdaIN in conjunction with it. In the first step, a model is trained solely on the source domain using both input images and labels. In the second step, the model is initialized with the weights from the first step and is trained with images from the target domain in an unsupervised manner, as well as with source images in a supervised manner. 
Two losses are employed. 
The first is a domain confusion loss (class-wise adversarial loss) between the features of the source and target domain. 
The second loss is a cross entropy pixel-wise loss between the output of the model on source inputs and source labels.
In the third step, a pseudo labeling approach is performed. Using the model trained in the second step, pseudo labels are generated for the target images.  The network is retrained with these pseudo labels on the target domain. 

pAdaIN is applied during all three stages with the same $p=0.01$ as in all other benchmarks. In the second loss of the second step, a slight modification is made to the training procedure of~\cite{Wang2020FADA}.
In the original setting, only source domain images are used in a given batch. 
Instead, when using pAdaIN, the inputs from the target domain are concatenated to the batch of source domain images. Thus, when applying the forward step on inputs from the source domain, pAdaIN mixes statistics from the target domain to the source domain.  
Note that we do not modify the target domains' image features in the forward pass, since we want to adapt to the target domain itself.
Also, while normally pAdaIN is applied on a larger batch size, due to GPU memory constraints, a batch-size of two is used, with one image from each domain.

We evaluate our method on the GTAV dataset~\cite{Richter_2016_ECCV_gtav} as our source domain and Cityscapes dataset as the target domain~\cite{Cordts2016Cityscapes} and use the official implementation and training scheme of FADA~\cite{Wang2020FADA}. GTAV is a synthetic dataset with $24,966$ urban scene images sourced from a video game rendering engine Grand Theft Auto V and Cityscapes is a real world urban scene dataset with $2975$ training images and $500$ validation images. 
An mIOU metric is used for evaluation.
For a fair comparison to previous methods, we evaluate each image in a single scale.
As can be seen in Tab.~\ref{tab:gta5_results}, our method improves the state of the art when applied in conjunction with FADA~\cite{Wang2020FADA}, achieving a gap of $2.3$ mIOU. There is also a gap  of $1.7$ when adding pAdaIN to a model trained only for the first phase, i.e., without access to target domain images.  
We notice that the greatest improvements occur on large objects, such as buses, trains, trucks, sidewalks and walls and a degradation in the sky category.

\subsection{Multi-domain Generalization}
\label{sec:generalization}

A more restrictive setting than that of domain adaptation is that of \textit{domain generalization}, in which the unlabeled target data is not available during training. The ``source only'' setting for the GTAV to Cityscapes experiment in Tab.~\ref{tab:gta5_results}, described above, is one instance of this problem. 

We evaluate our method for domain generalization on the PACS dataset~\cite{Li2017dg}. It consists of four domains: photo, art, cartoon, and sketch. We follow the multi-source evaluation protocol of \cite{carlucci2019domain}, training on three out of the four domains and evaluating on the fourth. We compare with the latest domain generalization methods. For the baseline method comparison, we simply train a network on the source data, without further modifications. Our models
are trained with SGD, over 30 epochs, batch size 128. 
The learning rate is set to $0.001$.
For the RSC~\cite{huang2020self} method, we also independently run the method using the open-source implementation as published by the authors,  using the default configuration (\url{https://github.com/DeLightCMU/RSC}). 

Tab.~\ref{tab:result_multi_dource_dg} shows the results of our method against that of baseline methods. As can be seen, when trained with pAdaIN, our method, on average, beats all baseline methods on both ResNet18 and ResNet50, except when considered against the reported values of RSC~\cite{huang2020self}. We note that our method outperforms our independently reproduced results of RSC, which follows the official open source implementation.
Our results improve over the baseline method especially in the sketch domain. 
Sketch images consist mostly of the outlines of the objects with no texture. This indicates, as previously shown, that
performance based on global cues, such as the object's outline (shape), is enhanced in this case.

\begin{table}
\begin{tabular}{lccccc}
\toprule
Method  & Photo & Art & Cart & Sketch & Avg \\
\midrule
Baseline~\cite{carlucci2019domain}  & 95.98 & 77.87 & 74.86 & 70.17 & 79.72 \\
D-SAM~\cite{d2018domain}   & 95.30 & 77.33 & 72.43 & \textbf{77.83} & 80.72 \\
JiGen~\cite{carlucci2019domain} & 96.03 & 79.42 & 75.25 & 71.35 & 80.51   \\
MASF~\cite{dou2019domain}   & 94.99 &  80.29 & \textbf{77.17} & 71.69 & 81.03    \\
E-FCR~\cite{li2019episodic}  & 93.90 & 82.10 & 77.00 & 73.00  & 81.50  \\
MetaReg~\cite{balaji2018metareg} & 95.50 & \textbf{83.70} & 77.20 & 70.30  & 81.70 \\
I-Drop~\cite{shi2020informative}  & 96.11 & 80.27 & 76.54 & 76.38 & 82.32 \\
RSC$^{*}$~\cite{huang2020self} 
& 95.99 & 83.43 & 80.31  & 80.85 &  85.15  \\
RSC$^{**}$~\cite{huang2020self} 
& 94.10 & 78.90 & 76.88 & 76.81 & 81.67 \\
\textbf{Ours}   & \textbf{96.29} & 81.74 & 76.91 & 75.13 & \textbf{82.51}   \\
\midrule 
Baseline~\cite{carlucci2019domain}  & \textbf{97.66} & 86.20 & 78.70 & 70.63  & 83.29\\ %
MASF~\cite{dou2019domain}   & 95.01 & 82.89 & 80.49 & 72.29 & 82.67 \\ %
MetaReg~\cite{balaji2018metareg} & 97.60 & \textbf{87.20} & 79.20 & 70.30  & 83.60 \\ %
RSC$^{*}$~\cite{huang2020self} 
& 97.92 & 87.89 & 82.16 & 83.35 & 87.83  \\ %
RSC$^{**}$~\cite{huang2020self} 
& 93.72 & 81.38 & 80.14 & \textbf{82.31} & 84.38  \\ %
\textbf{Ours} & 97.17 & 85.82 & \textbf{81.06} & 77.37 & \textbf{85.36} \\ %
\bottomrule
\end{tabular}
\caption{Results on multi-source domain generalization on the PACS dataset. \textbf{Top:} ResNet18, \textbf{Bottom:} ResNet50. Highlighted are the best scores per category. We consider RSC~\cite{huang2020self} reproduced scores ($^{**}$) and not the reported ones ($^*$). See Sec.~\ref{sec:generalization}  for details. Cart stands for Cartoon.  }
\label{tab:result_multi_dource_dg}
\vspace{-0.3cm}
\end{table}

\begin{table*}
{\setlength\tabcolsep{1.3318pt}%
\begin{tabular}{l l l c c  c c c  c c c c  c c c  c  c c c c@{}}
\toprule
 Dataset & Network & Architecture &E & mCE &  \multicolumn{3}{c}{Noise} & \multicolumn{4}{c}{Blur} & \multicolumn{4}{c}{Weather} & \multicolumn{4}{c}{Digital} \\
\cmidrule(lr){6-8}
\cmidrule(lr){9-12}
\cmidrule(lr){13-16}
\cmidrule(lr){17-20}
& & & & & \scriptsize{Gauss.}
    & \scriptsize{Shot} & \scriptsize{Impulse} & \scriptsize{Defocus} & \scriptsize{Glass} & \scriptsize{Motion} & \scriptsize{Zoom} & \scriptsize{Snow} & \scriptsize{Frost} & \scriptsize{Fog} & \scriptsize{Bright} & \scriptsize{Contrast} & \scriptsize{Elastic} & \scriptsize{Pixel} & \scriptsize{JPEG}\\ \hline
INet-C & Baseline  &ResNet50& 22.9 & 76.7  & 80  & 82  & 83  & 75  & 89  & 78  & 80  & 78  & 75  & 66  & 57  & 71  & 85  & 77  & 77 \\
INet-C & pAdaIN  & ResNet50& \textbf{22.3} & \textbf{72.8}  & \textbf{78}  & \textbf{79}  & \textbf{81}  & \textbf{70}  & \textbf{87}  & \textbf{74}  & \textbf{76}  & \textbf{74}  & \textbf{71}  & \textbf{64}  & \textbf{55}  & \textbf{65}  & \textbf{82}  & \textbf{66}  &\textbf{71} \\
\midrule
C100-C & Augmix~\cite{hendrycks2019augmix} & DenseNet-BC & 24.2 & 38.9  & 60  & 51  & 41  & 27  & 55  & 31  & 29  & 36  & 39  & 35  & 28  & 37  & 33  & 39  & 41 \\
C100-C & Augmix+pAdaIN &DenseNet-BC & \textbf{22.2} & \textbf{37.5}  & \textbf{58}  & \textbf{49}  & \textbf{40}  & \textbf{26}  & \textbf{54}  & \textbf{30}  & \textbf{28}  & \textbf{35}  & \textbf{38}  & \textbf{33}  & \textbf{25}  & \textbf{36}  & \textbf{32}  & \textbf{37}  &\textbf{40} \\
\midrule
C100-C & Augmix~\cite{hendrycks2019augmix} & ResNext-29 & 21.0 & 34.4  & \textbf{56}  & \textbf{48}  & 32  & 23  & \textbf{49}  & 27  & 25  & 32  & 35  & 32  & 24  & 32  & 30  & 34  & 37 \\
C100-C & Augmix+pAdaIN & ResNext-29  & \textbf{17.3} & \textbf{31.6}  & 58  & \textbf{48}  & \textbf{24}  & \textbf{20}  & 54  & \textbf{23}  & \textbf{21}  & \textbf{28}  & \textbf{30}  & \textbf{25}  & \textbf{19}  & \textbf{27}  & \textbf{27}  & \textbf{33}  &\textbf{36} \\
\bottomrule
\end{tabular}}
\caption{Clean Top-1 Error (E), Mean Corruption Error (mCE) and Corruption Error values of various corruptions. First, we consider an ImageNet trained ResNet50 model with or without pAdaIN, evaluated on \textsc{ImageNet-C}  (INet-C). Second, we consider DenseNet and ResNext models trained on CIFAR-100 either with Augmix alone or together with pAdaIN and evaluated on \textsc{CIFAR-100-C} (C100-C).
}\label{tab:result_imagenetc}

\end{table*}

\begin{table*}
\begin{tabular}{lcccccccc}
\toprule
& Baseline & Cutout & Mixup & CutMix &Auto- & Adversarial & Augmix & pAdaIN+  \\
& &  \cite{devries2017improved} & \cite{yun2019cutmix}  & \cite{yun2019cutmix} & Augment~\cite{cubuk2018autoaugment} & Training~\cite{madry2017towards} & \cite{hendrycks2019augmix}  & Augmix\\
\midrule
DenseNet-BC & 59.3 & 59.6 &  55.4 &  59.2 &  53.9 &  55.2  & 38.9 & \textbf{37.5} \\
ResNext-29 & 53.4 & 54.6 & 51.4 &  54.1 &  51.3  & 54.4 & 34.4 & \textbf{31.6} \\
\bottomrule

\end{tabular}
\caption{Classification error in comparison to state of the art baselines on CIFAR-100-C for ResNext~\cite{xie2017aggregated} and DenseNet~\cite{huang2017densely}. pAdaIN in conjunction with Augmix~\cite{hendrycks2019augmix} exceeds the state of the art. Baseline indicates a network trained on CIFAR-100 without any modifications. }
\label{tab:average_corruption}
\vspace{-0.3cm}
\end{table*}

\begin{table}
\begin{tabular}{l c c c c c c c c}
\toprule
  0  &  1    & 2       & 3   & 1-3 &  4 & 3-4 & all  \\
\midrule
76.1 & 75.9 &  76.1    &  76.5 & 76.4  & 77.5     & 78.1    & 77.8 \\
\bottomrule
\end{tabular}
\caption{Accuracy (bottom) on different block numbers (top) for which pAdaIN is applied on a ResNet18 trained on CIFAR100.}
\label{tab:cifar_classification_layer_ablation}
\vspace{-0.2cm}
\end{table}

\begin{table}
\begin{tabular}{l c c c c c c c c c}
\toprule
  $\mu(x_{i})$, $\sigma(x_{i})$ & Yes & Yes & No &  No & -\\
  $\mu(x_{\pi(i)})$, $\sigma(x_{\pi(i)})$ & No & Yes & Yes &  No & -\\ 
\midrule
Accuracy & $77.8$ & $77.6^*$  & $75.2$ & $75.1$ & $76.1$ \\
\bottomrule
\end{tabular}
\caption{
Accuracy for alternative backpropagation schemes for a ResNet18 trained on CIFAR100. Yes indicates backprop and No otherwise. We advocate for the leftmost scheme. Rightmost column is without using pAdaIn. *indicated unstable training.}
\label{tab:backprop_ablation}
\vspace{-0.5cm}
\end{table}

\subsection{Robustness Towards Corruptions}

Convolutional neural networks tend to be sensitive to small perturbations \cite{dodge2017study}. These small perturbations affect the statistics of the representation layers of the network. It is thus plausible that a model taught to be insensitive to statistical shifts of the feature space, such as our method, would be more robust towards corruptions.
To test this hypothesis, we evaluate our method against ImageNet-C and Cifar-100-C \cite{dodge2017study}, corrupted versions of ImageNet and CIFAR100.

First, we consider a ResNet50 model trained on ImageNet with or without pAdaIN. As can be seen in Tab.~\ref{tab:result_imagenetc}, our method improves upon the  baseline method trained without pAdaIN (with $p=0.01$).  Next, we consider pAdaIN in conjunction with AugMix~\cite{hendrycks2019augmix}, which is the current state of the art. As can be seen, combining pAdaIN with Augmix exceeds Augmix and is thus state of the art. For reference, average test error for additional methods designed for corruption are reported in Tab.~\ref{tab:average_corruption}.
Here we note that our smallest improvement is for the noise, blur, pixelated and JPEG corruptions, since these preserve the global statistics and have a tendency to modify the the fine details. Conversely, weather and contrast corruptions preserve texture and we, therefore, see an overall greater improvement for these categories.

\subsection{Ablation Analysis}
\label{sec:ablation}

To further evaluate different variants of pAdaIN, we consider a ResNet18 network trained on CIFAR100, as described in Sec.~\ref{sec:classification}. In Tab.~\ref{tab:cifar_classification_layer_ablation} we consider the effect of using pAdaIN on specific blocks of the ResNet18 network.  In all other experiments, we apply it to all layers. As can be seen, the effect of pAdaIN is most prominent when applied at the deeper blocks, specifically at blocks $3$ and $4$.

Next, we wish to understand the importance of using statistics, 
of the feature representation from  natural images.
Instead of swapping statistics between the feature representations of images, we swap the statistics of an image's feature representation with random statistics sampled from a normal distribution with zero mean and unit variance. We set the probability for this to happen at $p=0.01$, as in the default pAdaIN setting. We observed that as the model converged to minimal loss on the training set, the validation performance was very unstable, both in terms of loss and test accuracy. The overall accuracy is $57.3$, which is significantly lower. We believe this is due to the distribution shift from the statistics of natural images, happening with probability $p$.

As discussed in Sec.~\ref{sec:motivation}, in our method, we regard $\mu(x_{\pi(i)})$ and $\sigma(x_{\pi(i)})$ as constants and do not backpropagate through them. As can be seen in Tab.~\ref{tab:backprop_ablation}, setting $\mu(x_{i})$ and $\sigma(x_{i})$ as constant results in a degradation in performance. 
Applying backpropagation through $\mu(x_{\pi(i)})$ and $\sigma(x_{\pi(i)})$ as well, results in unstable training.

In addition, we analyze the effect of applying a fixed permutation across all layers in contrast to uniformly drawing one at each step of pAdaIN independently in the forward pass. To this end, we consider a ResNet18 model trained on CIFAR100, this reduces the accuracy to $68.02$, compared to pAdaIN accuracy of $77.82$ and $76.13$ for the baseline model.

\noindent\textbf{Computational time.}\quad We measure the additional time incurred by incorporating pAdaIN into training. Using the same computational resources (4$\times$ Nvidia V100 GPUs) the training times for a ResNet50 on ImageNet for $300$ epochs is 108 hours with and without pAdaIN (p=0.01). Thus training with pAdaIN does not result in increase in time complexity.

\section{Conclusions}

While CNN image classifiers are extremely powerful, they are still reliant on global image statistics that are easy to manipulate without changing the image semantics. In this work, we make use of the normalization mechanism in order to remove the reliance on this bias. The method is probabilistic and has a parameter $p$ that controls the tradeoff between training on deliberately mismatched image statistics and employing the matching global statistics. Naturally, there is exploitable information in these statistics that can help in image recognition benchmarks.

Since the texture is often defined as image statistics, and since previous work has focused on removing the bias toward texture, it is important to make the distinction between texture and global image statistics. As our motivating example shows, texture patterns are largely invariant to changes in global image statistics, even if these occur simultaneously across multiple encoding channels.

Indeed, contrary to results of methods for correcting texture bias, we demonstrate that the increase in classification performance goes hand in hand with the increase in classification capabilities on texture datasets. We do not believe this to be a misinterpretation by previous work, since we tested the performance of selected texture bias removal methods on texture datasets and observed a decrease in performance. We, therefore, view the two effects as distinct. 

Despite this distinctiveness, both our and texture bias removal methods demonstrate an increase in the recognition ability in the face of domain shifts. As future work, we would like to explore combining the two approaches together.

\section*{Acknowledgment}
This project has received funding from the European Research Council (ERC) under the European Unions Horizon 2020 research and innovation programme (grant ERC CoG 725974).

In addition, we would like to sincerely thank Sharon Fogel for her
substantial advisory support to this work.


{\small
\bibliographystyle{ieee_fullname}
\bibliography{padain}
}

\end{document}